\documentclass[final,3p,times,twocolumn]{elsarticle}

\usepackage{amssymb}
\usepackage{amsmath}

\usepackage{times}
\usepackage{helvet}
\usepackage{courier}
\usepackage{xcolor}

\usepackage[table]{xcolor} 
\usepackage{booktabs}

\newcommand{\eqautoref}[1]{\hyperref[#1]{Eq.~(\ref*{#1})}}

\journal{Pattern Recognization}

\begin{document}

\begin{frontmatter}


\ead{jfan@hdu.edu.cn (J. Fan), huby@hdu.edu.cn (B. Hu), lxguang@szpu.edu.cn (X. Li), yyx@hdu.edu.cn (Y. Yang), jingzhang.cv@gmail.com (J. Zhang)}

\title{FMGS-Avatar: Mesh-Guided 2D Gaussian Splatting with Foundation Model Priors for 3D Monocular Avatar Reconstruction}


\author[u1]{Jinlong Fan} 
\author[u1]{Bingyu Hu}
\author[u3]{Xingguang Li}
\author[u1]{Yuxiang Yang}
\author[u2]{Jing Zhang}
\affiliation[u1]{organization={HangZhou Dianzi University},
            addressline={}, 
            city={HangZhou},
            postcode={}, 
            state={Zhejiang},
            country={China}}
\affiliation[u3]{organization={Shenzhen Polytechnic University},
            addressline={}, 
            city={ShenZhen},
            postcode={}, 
            state={Guangdong},
            country={China}}         
\affiliation[u2]{organization={WuHan University},
            addressline={}, 
            city={WuHan},
            postcode={}, 
            state={Hubei},
            country={China}}

\begin{abstract}
Reconstructing high-fidelity animatable human avatars from monocular videos remains challenging due to insufficient geometric information in single-view observations. While recent 3D Gaussian Splatting methods have shown promise, they struggle with surface detail preservation due to the volumetric nature of 3D Gaussian primitives. To address both the representation limitations and information scarcity, we propose a novel method, \textbf{FMGS-Avatar}, that integrates two key innovations. First, we introduce Mesh-Guided 2D Gaussian Splatting, where 2D Gaussian primitives are attached directly to template mesh faces with constrained position, rotation, and movement, enabling superior surface alignment and geometric detail preservation. Second, we leverage foundation models trained on large-scale datasets, such as Sapiens, to complement the limited visual cues from monocular videos. However, when distilling multi-modal prior knowledge from foundation models, conflicting optimization objectives can emerge as different modalities exhibit distinct parameter sensitivities. We address this through a coordinated training strategy with selective gradient isolation, enabling each loss component to optimize its relevant parameters without interference. Through this combination of enhanced representation and coordinated information distillation, our approach significantly advances 3D monocular human avatar reconstruction. Experimental evaluation demonstrates superior reconstruction quality compared to existing methods, with notable gains in geometric accuracy and appearance fidelity while providing rich semantic information. Additionally, the distilled prior knowledge within a shared canonical space naturally enables spatially and temporally consistent rendering under novel views and poses.
\end{abstract}




\end{frontmatter}



\section{Introduction}
High-fidelity, animatable digital avatar creation has become increasingly important for applications ranging from entertainment and healthcare to AR/VR and interactive simulations. Traditional Motion Capture (MoCap) approaches, while capable of producing high-quality results, require expensive equipment~\cite{loper2014mosh,mahmood2019amass} or controlled studio environments~\cite{joo2015panoptic}, limiting their accessibility. The development of methods that can create digital avatars from readily available monocular RGB videos would significantly democratize this technology.

Recent advances in neural rendering have opened new possibilities for digital human reconstruction from monocular videos. Neural Radiance Field (NeRF)~\cite{mildenhall2021nerf} based approaches have demonstrated photorealistic rendering capabilities, though their computational requirements often limit real-time applications~\cite{yu2023monohuman,weng2022humannerf,jiang2022selfrecon,peng2021neural,zhou2024animatable}. 3D Gaussian Splatting (3DGS)~\cite{kerbl20233d} has emerged as an attractive alternative, offering efficient rendering while maintaining high visual quality. Methods such as Animatable 3D Gaussians~\cite{liu2024animatable}, GaussianAvatar~\cite{hu2024gaussianavatar}, and GART~\cite{lei2024gart} have shown promising progress in combining efficient rendering with realistic avatar modeling.


However, monocular avatar reconstruction faces two fundamental, interconnected challenges: geometric ambiguity from single-view data and the limitations of existing representations. While recent works have explored attaching 3D Gaussian primitives to a mesh template (e.g., GoMAvatar~\cite{wen2024gomavatar}), the volumetric nature of 3D Gaussians is suboptimal for representing surfaces, often leading to noisy geometry or depth ambiguity. Furthermore, while foundation models, such as DINOv2~\cite{oquab2023dinov2}, SAM~\cite{kirillov2023segment}, and Sapiens~\cite{khirodkar2024sapiens}, can offer rich 2D priors (depth, normals, semantics) to alleviate geometric ambiguity, systematically distilling the multi-modal knowledge introduces a critical, unaddressed problem: optimization conflicts, where supervisory signals from different modalities compete and interfere with each other during training.



To address these challenges, we propose \textbf{FMGS-Avatar}, a novel method that leverages \textbf{F}oundation \textbf{M}odel priors and \textbf{M}esh-\textbf{G}uided 2D \textbf{G}aussian \textbf{S}platting to assist monocular human avatar reconstruction through systematic knowledge distillation. Rather than focusing solely on geometric or appearance enhancement, our approach distills comprehensive 2D knowledge, including semantic understanding, depth information, and surface normals, into 3D human avatars, aiming to improve both geometric and appearance quality while providing semantic annotations.

First, we propose Mesh-Guided 2D Gaussian Splatting, a representation inherently suited for surfaces. Unlike volumetric 3D Gaussians-based methods, our approach employs 2D Gaussian Splatting (2DGS)~\cite{huang20242d} as the core representation and takes the 2D primitives as surfels, naturally aligning with the surface manifold and providing a more geometrically faithful representation. This Mesh-Guided 2DGS design choice aims to improve surface alignment while maintaining the computational efficiency of 2DGS. 

Second, we develop a method to systematically distill priors from multiple modalities, but more importantly, we introduce a Coordinated Training Strategy to resolve the inherent optimization conflicts. This strategy, featuring selective gradient stopping, is a core architectural innovation that enables the stable fusion of competing losses (e.g., depth loss affecting position, normal loss affecting orientation). It transforms the use of foundation models from simple "external supervision" into a deeply integrated and coherent learning process.  


The resulting avatar representation, enhanced with distilled 2D knowledge, can be rendered under novel views and poses, naturally maintaining spatial and temporal consistency through the shared canonical space. Our experimental evaluation suggests that this approach achieves improved reconstruction quality compared to existing methods. Our main contributions include:
\begin{itemize}
    \item \textbf{A Synergistic Framework for Knowledge Distillation}: We present a unified approach where a surface-centric representation (Mesh-Guided 2DGS) and a conflict-aware training strategy (Coordinated Training) work in concert to enable the systematic and stable distillation of comprehensive knowledge (geometry, semantics) from 2D foundation models into a 3D avatar. Our method is designed to be extensible for incorporating additional 2D priors as foundation models continue to advance.
    \item \textbf{Mesh-Guided 2D Gaussian Splatting}: We demonstrate the superiority of constraining 2D Gaussian primitives to a template mesh through explicit position, rotation, and movement constraints for surface modeling, achieving better geometric fidelity and alignment compared to methods that use volumetric 3D Gaussians.
    \item \textbf{Coordinated Training Strategy}: We introduce a coordinated training strategy that addresses multi-modal optimization conflicts through selective gradient isolation, enabling each loss component to focus on its most relevant parameters while preventing mutual interference. This approach ensures coherent learning across different Gaussian parameters and all representation components.
\end{itemize}

\begin{figure*}
    \centering
    \includegraphics[width=1\linewidth]{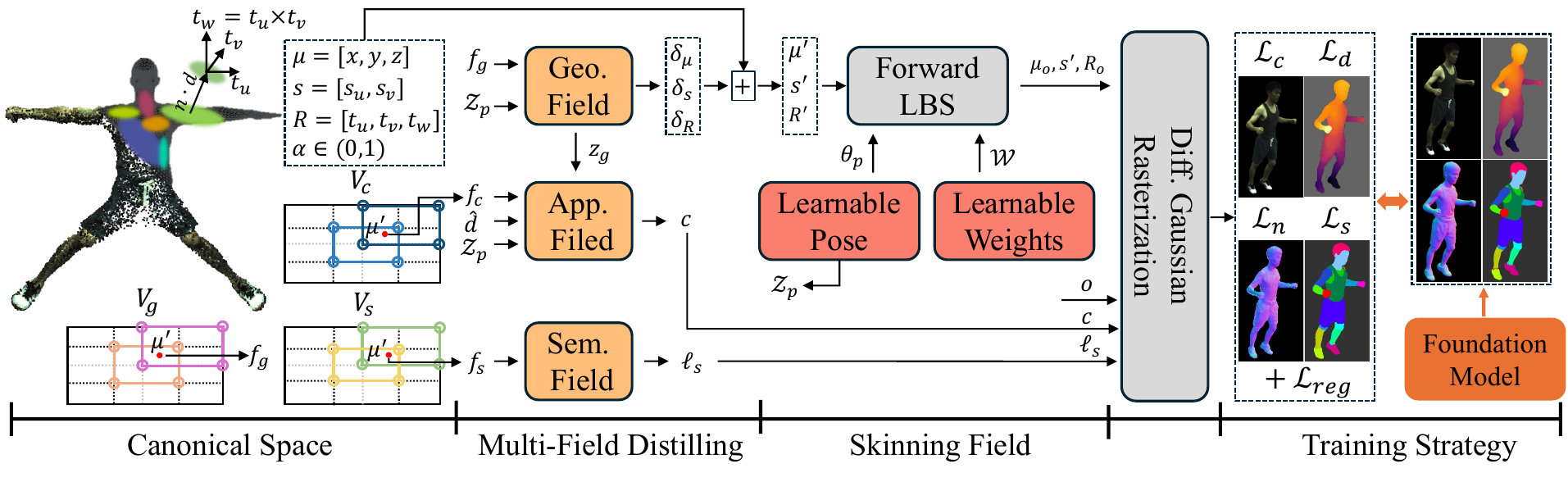}
    \caption{\textbf{Overview.} Our method distills foundation model priors to enhance monocular human avatar reconstruction through Mesh-Guided 2D Gaussian Splatting and multi-field knowledge distillation. (a) \textbf{Canonical Space Representation:} We constrain 2D Gaussian primitives to template mesh faces for superior surface alignment, while employing separate feature volumes $\mathcal{V}_g$, $\mathcal{V}_c$, and $\mathcal{V}_s$ to store geometry, appearance, and semantic properties, respectively. (b) \textbf{Multi-Field Distilling:} Based on sampled property features, we utilize corresponding property fields, including a pose-dependent geometry residual field, a view-dependent appearance field, and a semantic field, to capture distilled knowledge from foundation models. (c) \textbf{Skinning Field:} The canonical human representation with distilled knowledge is transformed to observation space through forward Linear Blend Skinning (LBS) using learnable pose parameters $\theta_p$ and predicted skinning weights $\mathcal{W}$. (d) \textbf{Training Strategy:} In addition to supervision losses on each rendered modality, we propose a coordinated training strategy to balance the multi-field optimization and resolve potential conflicts. This novel method enables high-quality monocular avatar reconstruction with enhanced geometric details and rich semantic properties through systematic 2D-to-3D knowledge transfer.}
    \label{fig:overview}
\end{figure*}

\section{Related Work}

\subsection{Monocular Human Avatar Reconstruction}
Early approaches for human avatar reconstruction relied on template-based methods that fit parametric models like SMPL to input observations~\cite{pavlakos2019expressive,kocabas2020vibe} but struggled with capturing clothing details. While NeRF-based methods~\cite{su2021nerf,jiang2022neuman,peng2021neural,xiu2023econ,pan2025litenerfavatar,huang2024efficient}, achieved photorealistic results, their slow rendering speeds have driven the community towards 3D Gaussian Splatting~\cite{kerbl20233d} for creating real-time animatable avatars.  

Recent 3DGS-based methods have explored various strategies~\cite{hu2024gaussianavatar,li2024animatable,hu2024gaussianavatar,shao2024splattingavatar,moreau2024human}. Some attach Gaussians to a template mesh to enforce structural consistency, such as GoMAvatar~\cite{wen2024gomavatar} and GauHuman~\cite{hu2024gauhuman}. Others, like 3DGS-Avatar~\cite{qian20243dgs}, focus on learning deformable Gaussian fields. The latest advancements continue to push the boundaries of expressiveness and efficiency. For instance, ExAvatar~\cite{moon2024expressive} extends the representation to the full body, including face and hands, by leveraging the SMPL-X model, enabling more expressive animations. Other works tackle the more challenging task of single-image reconstruction; GUAVA~\cite{zhang2025guava} achieves rapid upper-body avatar creation, and AniGS~\cite{qiu2025anigs} focuses on generating animatable avatars from a single, potentially inconsistent image. While these methods demonstrate remarkable progress, they primarily focus on appearance modeling and still inherit the limitations of using volumetric primitives to model thin surfaces. In contrast, our work proposes Mesh-Guided 2D Gaussians representation, providing a more natural and efficient representation for surface geometry. 



\subsection{Foundation Model Priors for 3D Reconstruction}
Foundation models have achieved remarkable success across diverse vision tasks, with general-purpose models like CLIP~\cite{radford2021learning}, DINOv2~\cite{oquab2023dinov2}, and SAM~\cite{kirillov2023segment} demonstrating exceptional zero-shot capabilities and robust feature representations. These models have been increasingly applied to 3D reconstruction, primarily for static scene reconstruction~\cite{kerr2023lerf,abou2025dino,zhou2024feature}. Concurrently, human-centric foundation models have rapidly emerged as a specialized domain~\cite{tang2025human}. Notably, Sapiens~\cite{khirodkar2024sapiens} represents a significant breakthrough, providing state-of-the-art performance across human pose estimation, depth prediction, surface normal estimation, and semantic parsing within a unified framework. Recent works have begun exploring the application of these human-centric foundation model priors to human avatar reconstruction. However, existing approaches predominantly leverage single-modal supervision in isolation. For example, StruGauAvatar~\cite{zhi2025strugauavatar} utilizes surface normals as pseudo ground truth. The systematic distillation of comprehensive multi-modal foundation model knowledge--encompassing depth, normals, and semantics--into a dynamic 3D human avatar remains largely unexplored. Furthermore, the inherent optimization conflicts that arise from simultaneously applying these diverse supervisory signals have not been adequately addressed.

\section{Preliminaries}

\paragraph{2D Gaussian Splatting}
 Unlike 3DGS, which uses 3D ellipsoids, 2DGS employs flat 2D Gaussian disks embedded in 3D space for scene representation. These primitives distribute densities within planar surfaces (surfels), enabling better surface alignment and improved geometry reconstruction compared to volumetric representations. Each 2D Gaussian primitive is characterized by its center point $\mathbf{\mu} \in \mathbb{R}^3$, opacity $\alpha \in \mathbb{R}$, view-dependent color $\mathbf{c} \in \mathbb{R}^3$ computed via spherical harmonics, scaling vector $\mathbf{s} = (s_u, s_v) \in \mathbb{R}^2$ controlling the 2D variance, and rotation matrix $\mathbf{R} \in \mathbb{R}^{3 \times 3}$. The rotation matrix $\mathbf{R} = [\mathbf{t}_u, \mathbf{t}_v, \mathbf{t}_w]$ consists of two orthogonal tangent vectors $\mathbf{t}_u, \mathbf{t}_v$ and the normal vector $\mathbf{t}_w = \mathbf{t}_u \times \mathbf{t}_v$ obtained through cross product. The 2D Gaussian is defined in a local tangent uv plane. For any point $\mathbf{u} = (u, v)$ in uv space, the Gaussian value is computed as $\mathcal{G}(\mathbf{u}) = \exp(-\frac{u^2 + v^2}{2})$. During rendering, 2DGS maps uv space to screen pixels through differentiable Gaussian rasterization:
\begin{equation}
    \mathbf{c}(\mathbf{x}) = \sum_i \mathbf{c}_i \alpha_i^{2D}  \prod_{j=1}^{i-1}(1 - \alpha_j^{2D} ).
    \label{eq:rendering}
\end{equation}



\section{Method}

Fig.~\ref{fig:overview} illustrates FMGS-Avatar for creating animatable 3D human avatars from monocular videos through systematic knowledge distillation from foundation models. Given a monocular video sequence $\{I^k\}_{k=1}^K$ with fitted SMPL parameters for each frame, including pose $\theta$, shape $\beta$, and template mesh $\mathcal{M}_c$, we first extract rich 2D priors including foreground masks $\bar{\mathcal{M}}$, pseudo depth $\bar{\mathcal{D}}$, surface normals $\bar{\mathcal{N}}$, and human parsing semantics $\bar{\mathcal{S}}$ using foundation model.


\subsection{Canonical Space Representation}  \label{sec:canonical}
\subsubsection{Mesh-Guided 2D Gaussian Splatting}  
%
The choice of 2D Gaussians over 3D is deliberate and critical for surface modeling. 3D Gaussians are volumetric ellipsoids. When representing a thin surface like cloth or skin, they must be made extremely flat, leading to training instabilities, or they retain volume, creating depth ambiguity and a "blurry" or "thickened" surface effect. 2D Gaussians, as planar surfels, are inherently surface-based primitives. This makes them a more efficient and geometrically faithful representation for avatar surfaces.

Given the SMPL template mesh $\mathcal{M}_c$ with 6,890 vertices, we first upsample it to 30,000 vertices to obtain a denser mesh $\mathcal{M}_c^{up} = \{\{v_i\}_{i=1}^V, \{f_j\}_{j=1}^F\}$ for enhanced surface detail representation. 
For each face $f_j$ on the upsampled mesh $\mathcal{M}_c^{up}$, we attach a corresponding 2D Gaussian primitive to establish explicit surface correspondence. This one-to-one mapping ensures that our representation can faithfully capture the underlying mesh topology while benefiting from the efficient rendering properties of Gaussian Splatting. For each 2D Gaussian primitive $k$, its position $\mathbf{\mu}_k$ is determined by the barycentric center of its corresponding face:
\begin{equation}  
\mathbf{\mu}_k = \frac{1}{3} \sum_{i=1}^{3} v_i,
\label{eq:barycentric_pos}  
\end{equation}  
where $v_i$ are the three vertices of face $f_k$. This constraint anchors each Gaussian to a specific mesh location, providing geometric stability and surface coherence.

The rotation matrix $\mathbf{R}_k = [\mathbf{t}_u, \mathbf{t}_v, \mathbf{t}_w]$ for each primitive is constructed such that the tangent vectors $(\mathbf{t}_u, \mathbf{t}_v)$ lie within the tangent plane of the mesh surface, while $\mathbf{t}_w$ aligns with the face normal $\mathbf{n}_k$. This orientation constraint ensures that 2D Gaussian primitives maintain proper surface alignment and follow the natural curvature of the human body.

Unlike conventional 3DGS methods that employ adaptive density control during optimization, our approach fixes the number of 2D Gaussians after mesh upsampling. This design choice prevents uncontrolled primitive proliferation while ensuring sufficient representation density through the systematic upsampling strategy. The fixed correspondence between mesh faces and Gaussians also facilitates consistent property feature learning across the avatar surface.



\subsubsection{Property Feature Sampling}  
Instead of using per-point features for encoding different avatar properties in canonical space, we employ separate HashGrid volumes~\cite{muller2022instant} for memory efficiency and multi-level feature fusion. We use independent volumes to store different property features: a geometry feature volume $\mathcal{V}_g$, an appearance feature volume $\mathcal{V}_c$, and a semantic feature volume $\mathcal{V}_s$. For each 2D Gaussian in canonical space, property features $(\mathbf{f}_g, \mathbf{f}_c, \mathbf{f}_s) \in \mathbb{R}^{32}$ are sampled from their corresponding feature volumes through trilinear interpolation. This design is extensible for incorporating additional foundation model properties by simply adding more feature volumes with minimal computational overhead, making our method adaptable to future foundation model advances.  

\subsection{Multi-Field Distilling}  \label{sec:distilling}
\subsubsection{Geometry Field}  
In canonical space, 2D Gaussian primitives constrained by the template mesh have limited representation ability for pose-dependent surface details. To address this, we introduce a pose-dependent geometry residual field to correct pose-related deformations. We formulate this field as a lightweight MLP, which takes geometry feature $\mathbf{f}_g$ and pose latent code $\mathcal{Z}_p$ as input:  
\begin{equation}  
(\delta d, \delta\mathbf{s}, \delta\mathbf{r}, \mathbf{z}_g) = \mathcal{F}_{\theta_g}(\mathbf{f}_g, \mathcal{Z}_p)  .
\label{eq:geometry_field}  
\end{equation}  
The pose latent code $\mathcal{Z}_p$ encodes SMPL pose and shape parameters $(\theta, \beta)$ using a hierarchical pose encoder~\cite{mihajlovic2021leap}, providing pose context for the observation space. The canonical Gaussians are corrected as:  
\begin{align}  
\mathbf{\mu}' &= \mathbf{\mu} + \mathbf{n} \cdot \delta d \label{eq:pos_correction},\\
\mathbf{s}' &= \mathbf{s} \cdot \exp(\delta\mathbf{s}) \label{eq:scale_correction},\\
\mathbf{R}' &= \mathbf{R} \cdot \exp([\delta\mathbf{r}]_{\times}), \label{eq:rot_correction}  
\end{align}  
where $\mathbf{n}$ is the surface normal. $\mathbf{n} \cdot \delta d$ ensures the position offset only moves along the normal direction, reducing the 3D movement freedom to 1D displacement. And $[\delta\mathbf{r}]_{\times}$ denotes the skew-symmetric matrix for rotation updates.

\subsubsection{Appearance Field}  
Conventional 3DGS methods use spherical harmonics for view-dependent color~\cite{wu20244d,yang2023real}, but in monocular settings, camera directions are limited and may not align with human pose variations. Similar to \cite{zhou2024animatable,qian20243dgs}, We canonicalize ray directions $\mathbf{d}$ from observation space to canonical space as $\hat{\mathbf{d}} = \mathbf{T}_{1:3,1:3}^{-1}\mathbf{d}$ using inverse rotation matrices from forward skinning (ref. to Sec.~\ref{sec:skinning}).  

Furthermore, local deformations such as clothing wrinkles depend on human pose, motivating pose conditioning for color prediction. Our appearance field takes sampled color feature $\mathbf{f}_c$, geometry-encoded feature $\mathbf{z}_g \in \mathbb{R}^{16}$ from geometry field, pose latent code $\mathcal{Z}_p \in \mathbb{R}^{16}$, and canonicalized viewing direction $\boldsymbol{\gamma}(\hat{\mathbf{d}})$ as input:  
\begin{equation}  
\mathbf{c} = \mathcal{F}_{\theta_c}(\mathbf{f}_c, \mathbf{z}_g, \mathcal{Z}_p, \boldsymbol{\gamma}(\hat{\mathbf{d}}))  .
\label{eq:appearance_field}  
\end{equation}  
Following \cite{qian20243dgs}, we use a compact MLP with one 64-dimensional hidden layer to prevent overfitting while maintaining sufficient representational capacity.

\subsubsection{Semantics Field}  
We leverage the Sapiens foundation model~\cite{khirodkar2024sapiens} to estimate the human parsing map with 28 semantic classes. To represent these semantics in our canonical space, we sample a feature vector $\mathbf{f}_s$ from the semantic feature volume $\mathcal{V}_s$. Directly interpreting these features as semantic logits (e.g., via a softmax) is suboptimal, as the feature volume stores a compressed, abstract representation rather than clean, class-specific logits.
Therefore, we employ a lightweight MLP, $\mathcal{F}_{\theta_s}$, which acts as a semantic decoder. This decoder learns a non-linear mapping from the sampled feature vector $\mathbf{f}_s$ to the final 28-dimensional semantic logits $\mathbf{l}_s$:
\begin{equation}  
\mathbf{l}_s = \mathcal{F}_{\theta_s}(\mathbf{f}_s)  .
\label{eq:semantic_field}  
\end{equation}  
Using a shared MLP decoder provides two key advantages: 1) It significantly increases the model's representational power, allowing it to learn complex boundaries between semantic regions. 2) It enhances spatial consistency by applying a single, coherent mapping function across the entire feature space, resulting in smoother and more reliable semantic maps. The rendered logits are then processed with softmax and argmax to obtain the final semantic map $I_s = \arg\max(\text{softmax}(\mathbf{l}_s))$.




\subsection{Skinning Field}  \label{sec:skinning}
Since mesh-guided 2D Gaussians corrected by the geometry residual field are no longer strictly on the template mesh, we have to diffuse the skinning weights defined on the mesh into 3D space. To that end, we learn a neural network $\mathcal{F}_{\theta_w}$ to predict the skinning weights $\mathcal{W} = \{w_b\}_{b=1}^{24}$ for any point in the canonical space.  

The input to this network are the corrected Gaussian positions $\mu^{\prime}$, which are first encoded into features using a multi-resolution hash encoding, which we denote as $H(\cdot)$. The network $\mathcal{F}_{\theta_w}$ is implemented as a 4-layer MLP with a hidden dimension of 128. This architecture takes the hash-encoded features as input and outputs a 24-dimensional vector corresponding to the influences of the SMPL joints. The final output is processed through a softmax layer to ensure the skinning weights sum to one ($\sum_{b=1}^{24} w_b = 1$). This process is formulated as: 
\begin{equation}  
\mathcal{W} = \text{softmax}(\mathcal{F}_{\theta_w}(H(\mu^{\prime}))).
\label{eq:skinning_field}  
\end{equation}  
We then transform the 2D Gaussian positions and rotations from canonical space to the observation space via forward Linear Blend Skinning (LBS):  
\begin{align}  
\mathbf{T} &= \sum_{b=1}^{24} w_b \mathbf{B}_b \label{eq:blend_transform}, \\
\mathbf{\mu}_o &= \mathbf{T} \mathbf{\mu}^{\prime} \label{eq:skinned_pos},\\
\mathbf{R}_o &= \mathbf{T}_{1:3,1:3} \mathbf{R}^{\prime}, \label{eq:skinned_rot}  
\end{align}  
where $\mathbf{B}_b$ represents the bone transformation matrices of human pose $\theta_p$, and $\mathbf{T}$ is the blended transformation matrix. Finally, images in different modalities are rendered via Eq.\ref{eq:rendering}.




\subsection{Training Objectives and Regularization}  \label{sec:loss}
Our training objective combines multiple loss terms to ensure high-quality reconstruction while effectively leveraging foundation model priors. 


\paragraph{Photometric Loss} 
We employ a combination of L1 and SSIM losses to ensure photometric consistency between rendered images $I$ and input frames $\bar{I}$:
\begin{equation}
\mathcal{L}_{\text{c}} = (1 - \lambda_{\text{ssim}}) \mathcal{L}_1(I,\bar{I}) + \lambda_{\text{ssim}} \mathcal{L}_{\text{SSIM}}(I,\bar{I}),
\label{eq:photo_loss}
\end{equation}
where $\lambda_{\text{ssim}} = 0.2$ balances the contribution of both terms. This combination captures both fine-grained pixel-level differences and perceptual image quality.

\paragraph{Mask Loss} 
To ensure accurate foreground segmentation, we apply binary cross-entropy loss on the rendered opacity mask $\mathcal{M}$ and ground truth mask $\bar{\mathcal{M}}$:
\begin{equation}
\mathcal{L}_{\text{m}} = \mathcal{L}_{\text{BCE}}(\mathcal{M},\bar{\mathcal{M}}).
\label{eq:mask_loss}
\end{equation}
\paragraph{Depth Supervision Loss} 
Since estimated single-view depth from foundation model exhibits scale ambiguity and may contain absolute depth errors, we employ an ordinal depth ranking loss that focuses on relative depth relationships rather than absolute values. Following~\cite{qingming2025modgs}, We define the ordinal indicator function as:
\begin{equation}
\mathcal{I}_{\text{ord}}(\mathcal{D}(x_1), \mathcal{D}(x_2)) = 
\begin{cases} 
+1, & \text{if } \mathcal{D}(x_1) > \mathcal{D}(x_2) \\ 
-1, & \text{if } \mathcal{D}(x_1) < \mathcal{D}(x_2) .
\end{cases}
\label{eq:ranking_function}
\end{equation}
The depth ranking loss is then formulated as:
\begin{equation}
\mathcal{L}_{\text{d}} = \left\|\tanh\left(\alpha (\mathcal{D}(x_1) - \mathcal{D}(x_2))\right) - \mathcal{I}_{\text{ord}}(\bar{\mathcal{D}}(x_1), \bar{\mathcal{D}}(x_2))\right\|_1,
\label{eq:depth_loss}
\end{equation}
where $\alpha = 10$ is a scaling factor, and we randomly sample pixel pairs $(x_1, x_2)$ during training to compute the ranking loss efficiently.

\paragraph{Surface Normal Loss} 
We supervise surface normal estimation using multiple consistency constraints. We use a self-consistency loss $\mathcal{L}_{sn}=\|\mathcal{N} - \hat{\mathcal{N}}\|_1$ between the rendered normals $\mathcal{N}$ and the normals derived from the rendered depth map, $\hat{\mathcal{N}}$. We also leverage a prior alignment loss $\mathcal{L}_{n}=1 - \mathcal{N} \cdot \bar{\mathcal{N}}$ to encourage consistency with the foundation model's predictions $\bar{\mathcal{N}}$. Additionally, to promote spatial smoothness on the rendered normal map, we apply a total variation (TV) regularization loss, $\mathcal{L}_{\text{tv}}$. The TV loss is defined as the sum of the absolute differences of neighboring pixel values in the normal map along the horizontal (x) and vertical (y) axes:  
\begin{equation}  
\mathcal{L}_{\text{tv}}(\mathcal{N}) = \sum_{i,j} \left( |\mathcal{N}_{i,j+1} - \mathcal{N}_{i,j}| + |\mathcal{N}_{i+1,j} - \mathcal{N}_{i,j}| \right).  
\label{eq:tv_loss}  
\end{equation}  
The complete surface normal loss is the sum of these components:  
\begin{equation}  
\mathcal{L}_{\text{norm}} = \mathcal{L}_{sn} + \mathcal{L}_n + \mathcal{L}_{\text{tv}}.  
\label{eq:normal_loss}  
\end{equation}  


\paragraph{Semantic Loss} 
For semantic supervision, we use standard cross-entropy loss between predicted semantic logits $\mathcal{S}$ and foundation model semantic maps $\bar{\mathcal{S}}$:
\begin{equation}
\mathcal{L}_{\text{s}} = \mathcal{L}_{\text{CE}}(\mathcal{S}, \bar{\mathcal{S}}).
\label{eq:semantic_loss}
\end{equation}
%

We also propose semantic-guided regularization that encourages feature consistency within semantic regions while allowing cross-region variations:
\begin{equation}
\mathcal{L}_{\text{reg}} = \frac{1}{|\mathcal{C}|} \sum_{c \in \mathcal{C}} \frac{1}{|\mathcal{G}_c|} \sum_{i,j \in \mathcal{G}_c} \|\mathbf{f}_i - \mathbf{f}_j\|_2^2,
\label{eq:semantic_reg}
\end{equation}
where $\mathcal{C}$ represents the set of semantic classes, $\mathcal{G}_c$ contains all Gaussians belonging to semantic class $c$, and $\mathbf{f}_i$ denotes sampled semantic features.

\paragraph{Other Regularization Terms} 
To prevent overfitting in the monocular setting and maintain geometric consistency, we also include $\mathcal{L}_{\text{skin}}$ regularizes skinning weights, and $\mathcal{L}_{\text{iso}}$ encourages as-rigid-as-possible deformation by preserving local distances between neighboring Gaussians.



\begin{figure}[!htbp]
    \centering
    \includegraphics[width=0.9\linewidth]{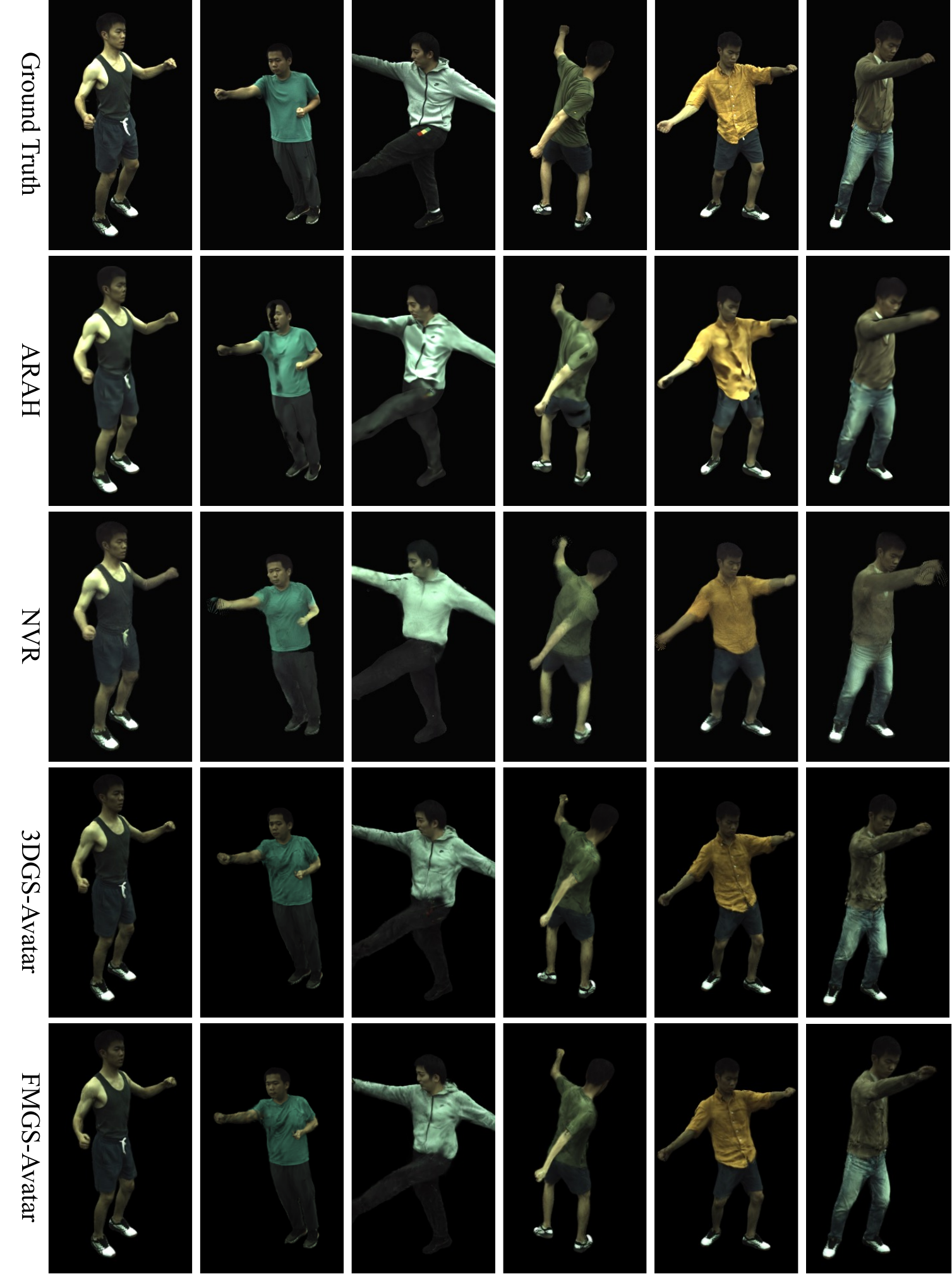}
    \caption{Qualitative results on ZJUMoCap dataset.}
    \label{fig:zjumocap}
\end{figure}

\paragraph{Total Loss Function} 
Our complete objective function combines all terms with carefully tuned weights:
\begin{align}
\mathcal{L} = &\mathcal{L}_{\text{c}} + \lambda_m \mathcal{L}_{\text{m}} + \lambda_d \mathcal{L}_{\text{d}} \nonumber \\
&+ \lambda_n \mathcal{L}_{\text{norm}} + \lambda_s \mathcal{L}_{\text{s}} + \lambda_{\text{reg}} \mathcal{L}_{\text{reg}} \nonumber \\
&+ \lambda_{\text{skin}} \mathcal{L}_{\text{skin}} + \lambda_{\text{iso}} \mathcal{L}_{\text{iso}},
\label{eq:total_loss}
\end{align}
%
where $\{\lambda_m, \lambda_{\text{skin}}, \lambda_{\text{iso}}\}$ follow established practices from~\cite{qian20243dgs}, while $\{\lambda_d, \lambda_n, \lambda_s, \lambda_{\text{reg}}\}$ are determined through empirical validation.

\subsection{Coordinated Training Strategy} 
Multi-field distillation faces conflicting optimization requirements where Depth estimation is primarily sensitive to Gaussian positions, surface normals depend critically on Gaussian orientations, and semantic information requires cluster consistency. To resolve these competing objectives targeting the same Gaussian parameters, we implement selective gradient stopping: (1) depth losses block gradients to Gaussian rotation parameters, focusing optimization on positional updates; (2) normal losses block gradients to Gaussian position parameters, concentrating on orientation refinement; and (3) semantic losses block gradients to both position and rotation parameters, directing optimization toward the semantic field while minimizing interference with geometric Gaussian parameters. This coordinated approach prevents parameter competition, enabling effective multi-modal knowledge distillation while maintaining geometric and semantic consistency for high-quality avatar reconstruction.

\section{Experiments}
In this section, we conduct comprehensive evaluations to demonstrate the effectiveness of our approach. We first compare our method with recent state-of-the-art methods for neural human reconstruction from monocular videos, including both NeRF-based (NeuralBody~\cite{peng2021neural}, Anim-NeRF~\cite{peng2021animatable}, ARAH~\cite{wang2022arah}, NVR~\cite{geng2023learning}, HumanNeRF~\cite{weng2022humannerf}) and 3DGS-based (GoMAvatar~\cite{wen2024gomavatar}, GauHuman~\cite{hu2024gauhuman}, 3DGS-Avatar~\cite{qian20243dgs}) approaches. Subsequently, we perform systematic ablation studies to validate the effectiveness of each designed component.




\begin{table}[!htbp]
\setlength{\fboxsep}{2pt}
\fontsize{9}{10}\selectfont
 \centering
 \setlength{\tabcolsep}{2pt}
 \begin{tabular}{ r|ccccc}
 \toprule
 Method & PSNR$\uparrow$ & SSIM$\uparrow$ & LPIPS$\downarrow$ & Train$\downarrow$ & FPS$\uparrow$ 
 \\ \hline
 NeuralBody~\cite{peng2021neural}  & 29.03 & 0.964 & 52.29 & 10h  & 1.5\\ 
 Anim-NeRF~\cite{peng2021animatable} & 29.17 & 0.961 & 51.98 & 13h & 1.1\\
 MonoHuman~\cite{yu2023monohuman} & 30.26 & \cellcolor{yellow}0.969 & 30.92 & 6h &  0.1 \\
 InstantAvatar~\cite{jiang2023instantavatar} & 29.73 & 0.938 & 64.41 & \cellcolor{yellow}5m & 4.2\\
 HumanNeRF~\cite{weng2022humannerf} & 30.24 & \cellcolor{yellow}0.969 & 33.38 & 10h & 0.3\\
 GoMAvatar~\cite{wen2024gomavatar} & 30.56 & 0.967 & 32.55 & 15h &  43 \\
 GauHuman~\cite{hu2024gauhuman} & \cellcolor{yellow}30.79 & 0.960 & 32.73 & \cellcolor{pink}1m &  \cellcolor{pink}180\\ 
 3DGS-Avatar~\cite{qian20243dgs}   & 30.61 & 0.965 & \cellcolor{yellow}30.28 & 17m &  50\\ \hline
 FMGS-Avatar & \cellcolor{pink}30.89 & \cellcolor{pink}0.972 & \cellcolor{pink}28.59 & 10m & \cellcolor{yellow}55\\
 \bottomrule
 \end{tabular}

  \caption{\textbf{Quantitative Results on ZJU-MoCap.} Cell color indicated \colorbox{pink}{Best} and \colorbox{yellow}{Second Best}.
  \label{tab:zjumocap}
  }
  
\end{table}
\begin{table*}[!htbp]
\resizebox{\textwidth}{!}{
\begin{tabular}{r|ccc|ccc|ccc|ccc}
\toprule
& \multicolumn{3}{c|}{male-3-casual} & \multicolumn{3}{c|}{male-4-casual} & \multicolumn{3}{c|}{female-3-casual} & \multicolumn{3}{c}{female-4-casual}      \\ 
& PSNR$\uparrow$ & SSIM$\uparrow$ & LPIPS$\downarrow$  & PSNR$\uparrow$ & SSIM$\uparrow$ & LPIPS$\downarrow$  & PSNR$\uparrow$ & SSIM$\uparrow$ & LPIPS$\downarrow$ & PSNR$\uparrow$ & SSIM$\uparrow$ & LPIPS$\downarrow$  \\ 
\hline
Anim-NeRF~\cite{peng2021animatable} & 23.17 & 0.9266 & 78.4  & 22.30 & 0.9235 & 91.1  & 22.37 & 0.9311 & 78.4  & 23.18 & 0.9292 & 68.7   \\
NeuralBody~\cite{peng2021neural} & 24.94	& 0.9428	& 32.6	& 24.71	& 0.9469 & 	42.3 & 23.87	& 0.9504 & 34.6 & 	24.37	& 0.9451	& 38.2   \\
Anim-3DGS~\cite{liu2024animatable} & 29.06	& 0.9704	& 26.4	& 26.16	& 0.9554	& 49.1	& 24.59	& 0.9535	& 39.9	& 27.26	& 0.9634	& 28.1   \\
InstantAvatar~\cite{jiang2023instantavatar} & 29.65	& \cellcolor{yellow}0.9731	& \cellcolor{yellow}19.2	& 27.97	& 0.9649	& 34.6	& 27.9	& 0.9722	& 24.9	& 28.92	& \cellcolor{yellow}0.9692	& \cellcolor{yellow}18.0   \\
3DGS-Avatar~\cite{qian20243dgs} & \cellcolor{yellow}30.57	& 0.9581	& 20.9	& \cellcolor{yellow}33.16	& \cellcolor{yellow}0.9678	& \cellcolor{pink}15.7	& \cellcolor{pink}34.28	& \cellcolor{yellow}0.9724	& \cellcolor{yellow}14.9	& \cellcolor{yellow}30.22	& 0.9653	& 23.1   \\ \hline

FMGS-Avatar    & \cellcolor{pink}30.92	& \cellcolor{pink}0.9816	& \cellcolor{pink}15.2	& \cellcolor{pink}33.78	& \cellcolor{pink}0.9753	& \cellcolor{yellow}23.7	& \cellcolor{yellow}33.89 &\cellcolor{pink}0.9854 & \cellcolor{pink}14.2 & \cellcolor{pink}30.95	& \cellcolor{pink}0.9873	& \cellcolor{pink}15.8 \\
\bottomrule
\end{tabular}
}

\caption{\textbf{Comparison on PeopleSnapshot Dataset.} }
\label{tab:ps}
\end{table*}


\begin{figure}[!htbp]
    \centering
    \includegraphics[width=0.8\linewidth]{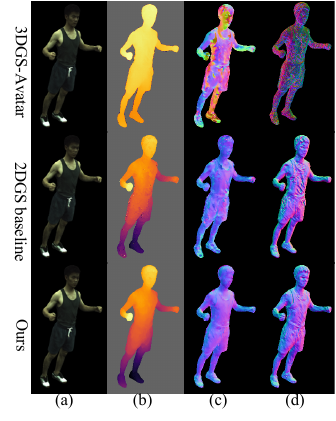}
    \caption{Comprehensive comparison of geometric reconstruction quality. From left to right: (a) reconstructed appearance, (b) depth map, (c) predicted normal, (d) surface normal derived from depth gradient. Our method demonstrates superior geometric fidelity.}
    \label{fig:geo}
\end{figure}

\subsection{Evaluation Datasets}

\paragraph{ZJU-MoCap Dataset} 
ZJU-MoCap dataset~\cite{peng2021neural} serves as our primary testbed for quantitative evaluation. We select six representative sequences (377, 386, 387, 392, 393, 394) from the ZJU-MoCap dataset and follow the standard training/test split established by HumanNeRF~\cite{weng2022humannerf}. The motion patterns in these sequences are repetitive and do not contain sufficient pose diversity for meaningful novel pose synthesis benchmarks. Therefore, we focus on evaluating novel view synthesis performance using standard metrics (PSNR/SSIM/LPIPS) and provide qualitative results for animation under out-of-distribution poses. Note that LPIPS values in all tables are scaled by 1000 for clarity.

\paragraph{PeopleSnapshot Dataset} 
We also conduct experiments on 4 sequences from PeopleSnapshot dataset~\cite{alldieck2018video}, which contains monocular videos of people rotating in front of a camera under controlled lighting conditions. We follow the data split protocol established by InstantAvatar~\cite{jiang2023instantavatar} and compare directly with their results for fair evaluation. For consistency, we use the SMPL poses optimized by AnimNeRF~\cite{peng2021animatable} without further refinement during our training process.





\subsection{Comparison with Baselines}

\paragraph{Quantitative Results}
Tab.\ref{tab:zjumocap} and \ref{tab:ps} present quantitative results for novel view synthesis on ZJU-MoCap and PeopleSnapshot datasets, respectively. Our method demonstrates superior rendering quality compared to both NeRF-based baselines and recent 3DGS-based state-of-the-art approaches across all evaluation metrics.

On ZJU-MoCap dataset, our method achieves state-of-the-art performance. Compared to the closest competitor, 3DGS-Avatar, we achieve consistent improvements across all metrics, while maintaining competitive rendering efficiency at 55 FPS. Our method shows particularly notable improvements over GoMAvatar, which shares a similar Gaussian-on-Mesh design philosophy but employs 3D Gaussians. We achieve substantial gains across all metrics. More importantly, our approach provides dramatic efficiency improvements with 90× faster training (10 minutes vs. 15 hours) while maintaining comparable inference speed (55 vs. 43 FPS).

On PeopleSnapshot dataset, our method consistently outperforms existing approaches across most sequences. Notably, we achieve the best performance on 3 out of 4 sequences, with particularly strong results on male-3-casual and female-4-casual, demonstrating the effectiveness of our foundation model distillation and mesh-guided representation across diverse subjects.

Our approach achieves significant training acceleration compared to traditional NeRF-based methods, requiring only 10 minutes versus hours for conventional approaches. While InstantAvatar achieves faster training (5 minutes), our method delivers substantially superior inference performance (55 FPS vs. 4.2 FPS), making it more suitable for real-time applications. Although some recent methods like GauHuman achieve faster training (1 minute) and higher inference speeds (180 FPS), our approach provides a better quality-efficiency trade-off, delivering superior reconstruction fidelity while maintaining practical rendering speeds for interactive applications.


\paragraph{Qualitative Analysis}
 Fig.\ref{fig:zjumocap} presents qualitative comparisons of novel view rendering results on ZJU-MoCap dataset. Our method produces significantly more detailed and geometrically consistent results compared to NeRF-based baselines while achieving comparable or superior quality to 3DGS-based state-of-the-art methods. NeRF-based methods exhibit characteristic limitations: ARAH~\cite{wang2022arah} shows notable artifacts on human body regions, particularly in areas with complex geometry, while NVR produces overly smooth surfaces that lack fine-grained details such as clothing wrinkles. In contrast, our approach effectively leverages distilled foundation model priors to resolve geometric ambiguities inherent in monocular reconstruction, resulting in enhanced surface details and more realistic appearance.
 
 
Fig.\ref{fig:geo} provides a comprehensive analysis of our method's geometric reconstruction capabilities through depth and surface normal visualizations. To systematically validate the effectiveness of our approach, we conduct comparative analysis using 3DGS-Avatar as the 3DGS baseline and an adapted 2DGS baseline where 3D Gaussians are replaced with conventional 2D Gaussians without mesh guidance.

The results demonstrate clear advantages of our approach across multiple geometric aspects. First, 2DGS primitives provide more consistent depth scaling and reasonable surface normal estimation compared to 3DGS, as the planar nature of 2D Gaussians could align with underlying surface geometry better. Second, our mesh-guided design further enhances this alignment by constraining primitives to template mesh faces, ensuring geometrically plausible surface reconstruction. Third, the integration of foundation model priors provides additional geometric cues that resolve ambiguities in monocular settings, leading to more accurate depth estimation and surface normal prediction.



\begin{figure}[!htbp]
    \centering
    \includegraphics[width=0.9\linewidth]{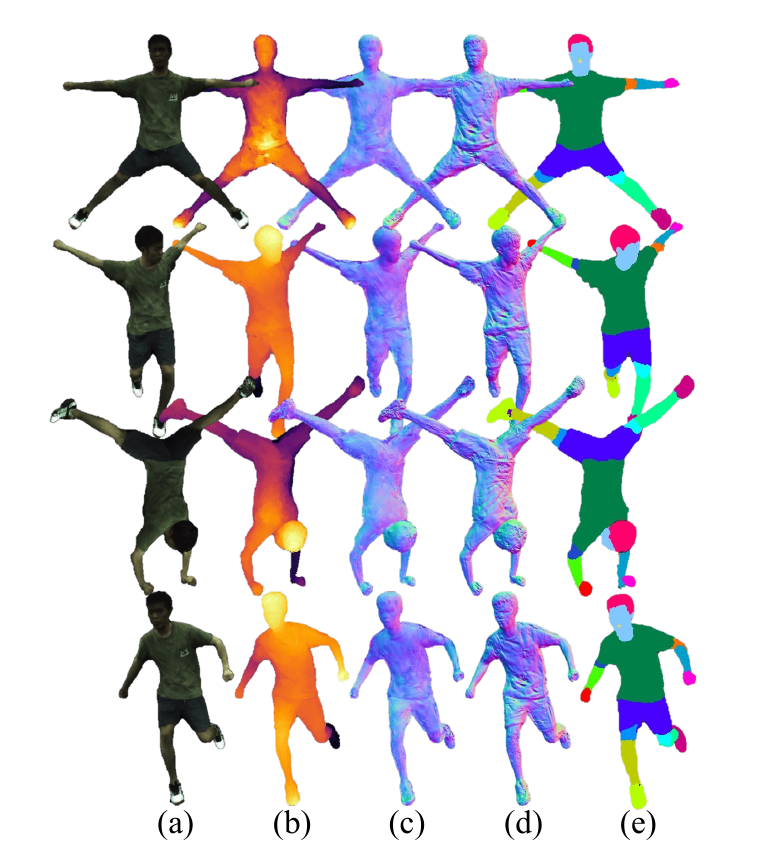}
    \caption{Multi-modal novel pose synthesis results showing: (a) RGB rendering, (b) depth map, (c) predicted normal, (d) surface normal derived from depth, (e) semantic segmentation. The first row shows the canonical rest pose, while subsequent rows demonstrate poses from AIST++~\cite{li2021ai} and AMASS~\cite{mahmood2019amass} sequences.}
    \label{fig:np}
\end{figure}

Fig.\ref{fig:np} demonstrates the rendered multi-modal results of Subject 392 driven by poses from AIST++~\cite{li2021ai} and AMASS~\cite{mahmood2019amass} sequences, showcasing our method's capability to generalize to out-of-distribution poses. This represents a significant advancement in 2D-to-3D knowledge transfer, where 2D semantic and geometric priors from foundation models become effectively drivable in 3D space. This drivable semantic information is particularly valuable for downstream applications such as virtual try-on, motion analysis, or avatar editing, expanding the practical utility of reconstructed avatars beyond basic animation and rendering.

\subsection{Ablation}

We conduct systematic ablation studies on Subject 377 from ZJU-MoCap dataset to validate the effectiveness of each proposed component. Tab.~\ref{tab:ablation} presents quantitative results demonstrating the progressive improvement achieved by each component.

\textbf{Foundation Model Supervision.} The baseline without foundation model supervision achieves the lowest performance (29.98 dB PSNR). Adding depth supervision ($\mathcal{L}_d$) provides a +0.33 dB improvement, demonstrating the value of geometric priors. Incorporating self-consistent normal loss ($\mathcal{L}_{sn}$) further enhances results by +0.42 dB, while normal supervision ($\mathcal{L}_n$) from foundation models contributes an additional +0.36 dB improvement. Finally, adding semantic supervision ($\mathcal{L}_s$) achieves the best performance (31.22 dB), validating the effectiveness of comprehensive multi-modal knowledge distillation.

\begin{figure}
    \centering
    \includegraphics[width=0.8\linewidth]{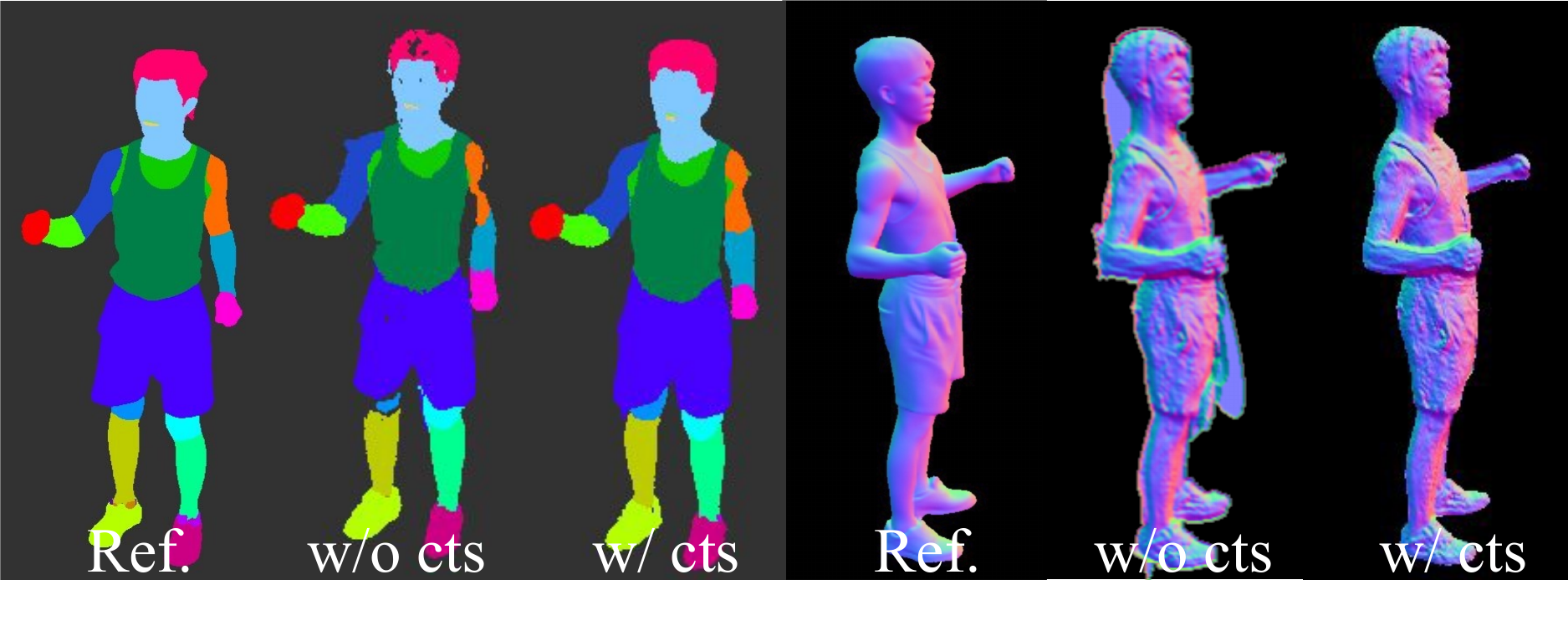}
    \caption{Effectiveness of Coordinated Training Strategy.}
    \label{fig:abl_cts}
\end{figure}

\textbf{Coordinated Training Strategy.} Fig.~\ref{fig:abl_cts} demonstrates the critical importance of our proposed Coordinated Training Strategy. Without selective gradient blocking, multi-modal losses compete for the same Gaussian parameters, leading to notable artifacts in both semantic and normal fields, evident as black holes in semantic maps and incorrect normals at the head and legs. Our coordinated approach effectively resolves these optimization conflicts, ensuring stable multi-field learning.

\begin{table}[htbp]
    \centering
    \resizebox{0.8\linewidth}{!}{
    \begin{tabular}{ccccccc}
        \hline
        $\mathcal{L}_{d}$ & $\mathcal{L}_{sn}$ & $\mathcal{L}_{n}$ & $\mathcal{L}_{s}$ & PSNR $\uparrow$ & SSIM $\uparrow$ & LPIPS $\downarrow$ \\
        \hline
                   &             &             &             & 29.98 & 0.938 & 29.53 \\
        \checkmark &             &             &             & 30.31 & 0.958 & 28.23 \\
        \checkmark & \checkmark  &             &             & 30.73 & 0.963 & 28.01 \\
        \checkmark & \checkmark  & \checkmark  &             & 31.09 & 0.975 & 27.42 \\
        \checkmark & \checkmark  & \checkmark  & \checkmark  & \textbf{31.22} & \textbf{0.978} & \textbf{26.53} \\
        \hline
    \end{tabular}
    }
    \caption{Quantitative results of ablation study.}
    \label{tab:ablation}
\end{table}

\section{Conclusion}
We propose FMGS-Avatar, which leverages mesh-guided 2D Gaussian Splatting with foundation model priors to enhance monocular human avatar reconstruction through systematic knowledge distillation. Our approach addresses three fundamental challenges: (1) information scarcity inherent in monocular observations through multi-modal foundation model distillation, (2) surface representation limitations of conventional 3D Gaussians through mesh-guided 2D Gaussians, and (3) optimization conflicts in multi-field learning through coordinated training strategies. 

Experimental results demonstrate that our method achieves state-of-the-art performance in both geometric accuracy and appearance fidelity while maintaining efficient training and rendering capabilities. The distilled 2D foundation model priors in canonical 3D space could be rendered under novel views and poses through spatially and temporally consistent avatar animation, significantly advancing the practical applicability of monocular avatar reconstruction.

Our method establishes a promising pathway for incorporating rapidly advancing foundation model capabilities into 3D human reconstruction, suggesting significant potential for future research in cross-modal knowledge transfer and neural avatar modeling.

\bibliographystyle{elsarticle-num} 
\bibliography{ref}

\end{document}